\documentclass[journal]{IEEEtran}
\usepackage{cite}
\newcommand{\authorcite}{{\textit{et}\,\textit{al}.}}
\newcommand{\argmax}[1]{\underset{#1}{\operatorname{max}}\;}

\usepackage{url}
\ifCLASSINFOpdf
   \usepackage[pdftex]{graphicx}
\else
\fi
\usepackage{amsmath,eqnarray,amsfonts,amssymb}
\usepackage{algorithmic}
\usepackage{array}
\ifCLASSOPTIONcompsoc
  \usepackage[caption=false,font=normalsize,labelfont=sf,textfont=sf]{subfig}
\else
 \usepackage[caption=false,font=footnotesize]{subfig}
\usepackage{textcomp}
\usepackage{multicol}
\usepackage{multirow}
\usepackage{xparse}
\usepackage{xcolor}
\usepackage{float}
\usepackage{adjustbox}
\usepackage{booktabs}
\begin{document}
%
\title{Multiple Instance-Based Video Anomaly Detection using Deep Temporal Encoding-Decoding}
%
%
%

\author{Ammar Mansoor Kamoona,~\IEEEmembership{Student Member,~IEEE,}
        Amirali Khodadadian Gostar, Alireza Bab-Hadiashar,~\IEEEmembership{Member,~IEEE,}
        Reza Hoseinnezhad,~\IEEEmembership{Fellow,~IEEE}
\thanks{A. M. Kammona with the Department
of Electrical and Computer Engineering, RMIT University, Melbourne, VIC, Australia.}
\thanks{J. Doe and J. Doe are with Anonymous University.}
\thanks{Manuscript received April 19, 2005; revised August 26, 2015.}}

\maketitle

\begin{abstract}
In this paper, we propose a weakly supervised deep temporal encoding-decoding solution for anomaly detection in surveillance videos using multiple instance learning. The proposed approach uses both abnormal and normal video clips during the training phase which is developed in the multiple instance framework where we treat the video as a bag and video clips as instances in the bag. Our main contribution lies in the proposed novel approach to consider temporal relations between video instances. We deal with video instances (clips) as sequential visual data rather than a set of independent instances. We employ a deep temporal encoding-decoding network that is designed to capture spatio-temporal evolution of video instances over time. 
We also propose a new loss function that maximizes the mean distance between normal and abnormal instance predictions. The new loss function ensures a low false alarm rate which is very crucial in practical surveillance application.The proposed temporal encoding-decoding approach with modified loss is benchmarked against the state of the art in simulation studies. The results show that the proposed method performs similar to or better than the state-of-the-art solutions for anomaly detection in video surveillance applications and achieve state of the art false alarm rate on UCF-crime dataset. 
\end{abstract}


%
\IEEEpeerreviewmaketitle

\section{Introduction}
\begin{figure*}
	\centering
	\includegraphics[width=\textwidth]{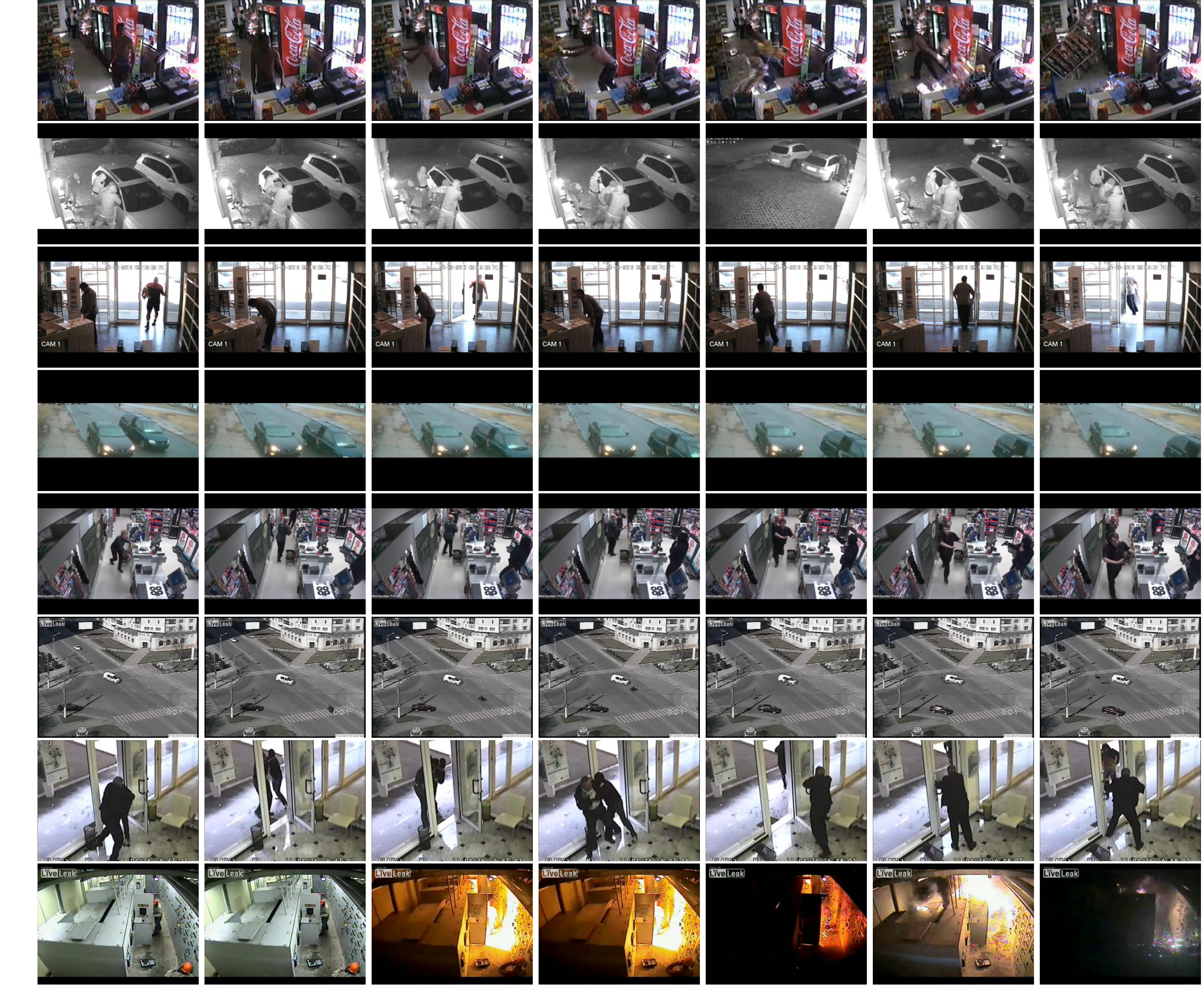}
	\small
	\put(-510,395){A}
	\put(-510,340){B}
	\put(-510,290){C}
	\put(-510,240){D}
	\put(-510,185){E}
	\put(-510,130){F}
	\put(-510,75){G}
	\put(-510,20){H}
	
	\caption{Different abnormal examples for different surveillance videos taken from UCF-crime dataset~\cite{sultani2018real}. (A): Vandalism; (B): Stealing; (C): Shoplifting; (D): Shooting; (E): Robbery; (F): Road Accident; (G): Robbery; (H): Explosion.}
	\label{Fig:anomaly_examples_1}
\end{figure*}
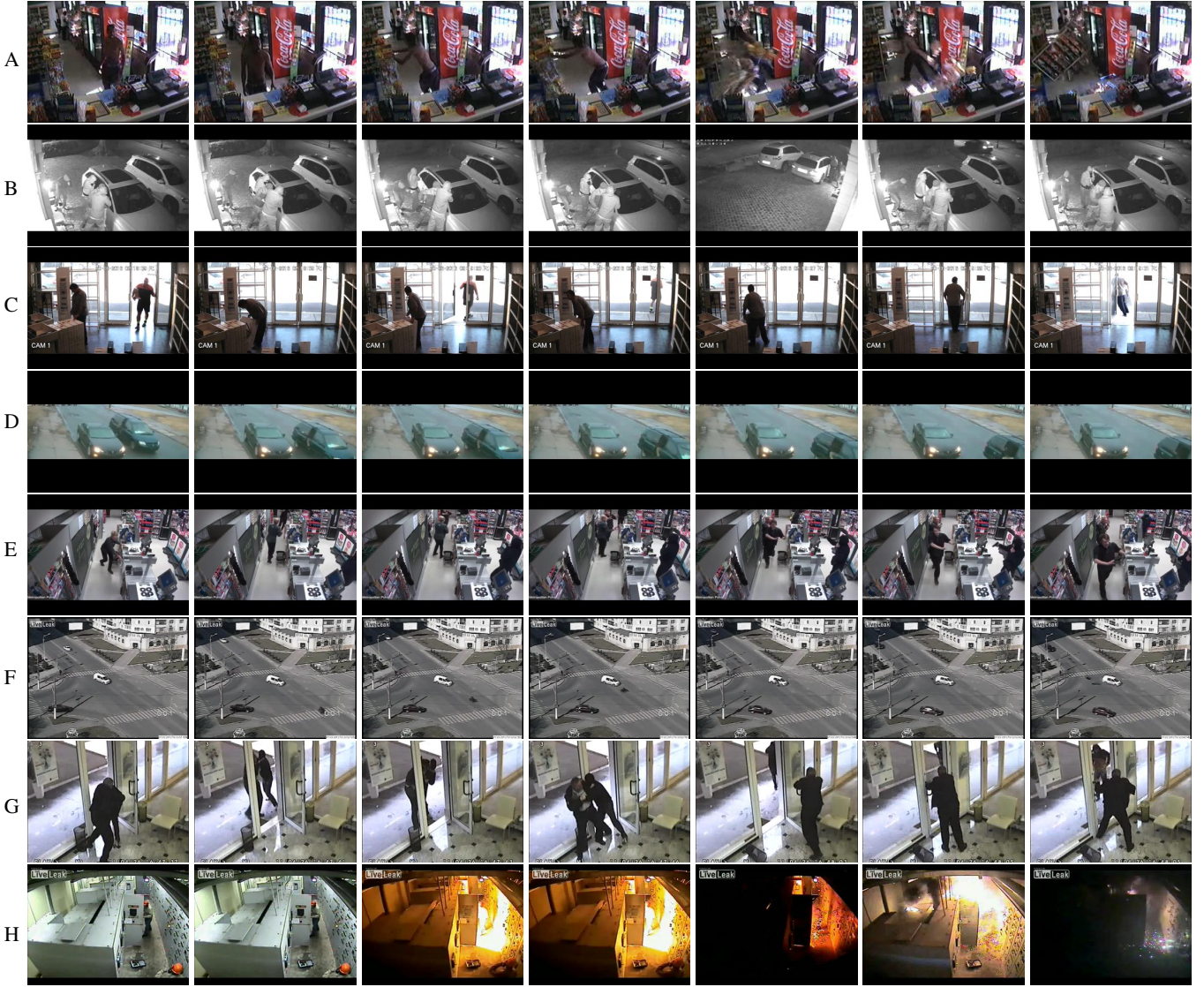

\IEEEPARstart{V}{ideo} anomaly detection is defined as the process of detecting the occurrence of ``abnormal" events in video clips that differ from previously defined ``normal" clips. Automatic detection of anomalies in video has gained a significant attention in the past few years~\cite{Abati_2019_CVPR,del2016discriminative,ionescu2019object,mehran2009abnormal,ravanbakhsh2017abnormal}. This is mainly due to the difficulty of manual processing (requires extensive manpower) of the abundant visual information generated by surveillance cameras. From security perspective, the detection of events such as stealing, fighting and shoplifting is of particular interest. Examples of such  activities are shown in Figure~\ref{Fig:anomaly_examples_1}.\\

	
Unsupervised anomaly detection is commonly formulated for detection of \textit{rare} abnormal events in which only frequently occurring behaviour of normal samples is used in the training. The most common approach is to treat abnormal events as \textit{outliers} to a model that is trained using normal videos~\cite{xia2015learning,liu2018future,luo2017revisit,hinami2017joint,sabokrou2017deep,smeureanu2017deep}. Unsupervised anomaly detection is usually performed either by using handcrafted features followed by feature learning or via development of an end-to-end deep network. The earlier approaches for anomaly detection commonly involved extraction of trajectory features to make use of its ability to describe the dynamics of moving objects~\cite{cai2015trajectory,piciarelli2008trajectory}. In addition, different spatio-temporal handcrafted features such as color, texture and optical flow have been used for anomaly detection~\cite{cong2011sparse,kratz2009anomaly,lu2013abnormal}. However, due to illumination changes, scale and deformation, these features do not generalize well for large scale video analysis. Therefore, unsupervised deep learning has been used for feature extraction and model learning~\cite{Abati_2019_CVPR}.

The aforementioned methods are based on normality deviation. However, Chandola\,\authorcite~\cite{chandola2009anomaly} showed that it is ambiguous to define a boundary between normal and abnormal, mostly due to the definition of normal events that can not take into account all possible normal patterns or behaviors. As a result, a any new occurrence of normal event may also deviate from the trained model and cause a false alarm~\cite{hasan2016learning}. Recently, a weakly supervised learning~\cite{sultani2018real,Zhong_2019_CVPR,zaheer2020claws} has gained a popularity which leverages the aforementioned problem by using of both normal and abnormal videos. A video is labeled as normal if all the video frames are normal and abnormal when video frames has normal and abnormal frames. The main reason for formalizing the problem within weak supervision is due to the lack of temporal labeling of abnormal videos.

Sultani\,\authorcite~\cite{sultani2018real} proposed to tackle the weakly supervised problem within the multiple instance learning framework where the bag (video) label is available and a model is trained to infer the instance label. They employed multiple instance hinge loss function and designed a network that processes video clips (independently from each other). Another weakly supervised learning approach is by training classifier under noisy labels~\cite{Zhong_2019_CVPR}. The noisy labels refer to normal segments in the anomalous video.

In this paper, we propose a new solution that hierarchically captures low, intermediate and high level temporal and spatial information. The contribution of this work is two-fold:
(i) we propose a novel solution for anomaly detection in videos that uses a temporal encoding network to capture the temporal and spatial information of video instances, and
(ii) the formulation of this solution is implemented within the weakly supervised multiple instance framework where we propose a loss function that has a smoother instances to bag mapping than its counterpart and penalize the false alarm. In addition, it achieves a competitive results compared to the state-of-the-art.

\section{Background}
In this section, a brief review of the most recent works on video-based anomaly detection methods is presented. Generally, the common approach for visual anomaly detection is based on extracting handcrafted or deep representation features, and model learning. In this approach, anomaly detection is usually formulated as an outlier detection problem.

Tracking-based anomaly detection methods were the earlier approaches used for dynamic feature extraction to model the normal pattern of movements of objects of the interest~\cite{piciarelli2008trajectory,cai2015trajectory}. In~\cite{wang2006learning}, anomaly detection based on semantic scene trajectory clustering was proposed. In this method, first blob detection is performed to detect the object of interest. Then feature extraction is performed using spatial and velocity information. These features are then clustered based on their similarities (in term of the size of the objects and their velocities) and the result is used to form trajectories. Hu\,\authorcite~\cite{hu2006system} proposed another anomaly detection technique using trajectory clustering of motion patterns. First, foreground pixels are detected using background subtraction method, then the features are clustered using the Fuzzy K-mean clustering algorithm. Anomaly detection is then carried out by comparing the learned motion pattern probability distribution obtained from the trajectories. In general, tracking-based methods are not robust enough for complex video scene analysis since they involve different complex steps such as object detection, data association and tracking, and any failure in these steps causes a failure in the anomaly detection system.

Due to the limitations of tracking-based methods, handcrafted spatio-temporal features have been employed to model the motion pattern for anomaly detection. The most straightforward approach is to extract low-level appearance features and motion cues such as color, texture and optical flow and use them to model motion activity patterns~\cite{cong2011sparse,lu2013abnormal,kratz2009anomaly,adam2008robust,mehran2009abnormal}. Mehran\,\authorcite~\cite{mehran2009abnormal} proposed to use social force model combined with optical flow features to learn normal pattern for global motion and any deviation from this model (with low probability) is considered an anomaly. Zhao\,\authorcite~\cite{zhao2015abnormal} used the Spatial Temporal Interest Point (STIP) detector to detect the region of interest and then Histogram of Gradient (HOG) as an appearance feature descriptor and Histogram of Optical Flow (HOF) as motion feature descriptor were used to detect abnormal activity in videos. Mahadevan\,\authorcite~\cite{mahadevan2010anomaly} tackled the problem of spatial and temporal anomaly detection in crowded scenes by jointly modeling a mixture of appearance and dynamics. Spatial anomalies are detected using discriminant saliency, while temporal anomalies are detected as an event with low probability.

Recently, unsupervised deep learning using autoencoder network has been widely used for latent features representation and anomaly detection~\cite{xu2015learning,hasan2016learning,Abati_2019_CVPR,nguyen2019anomaly}. Hasan\,\authorcite~\cite{hasan2016learning} used the reconstruction error of the fully connected convolution autoencoder as an anomaly score. Xu\,\authorcite~\cite{xu2015learning} proposed a rich and descriptive motion and appearance feature representation using a stacked autoencoder. Anomaly detection is performed based on anomaly score calculated using multiple one class SVM on the learned feature representation. Nguyen~and~Meunier~\cite{nguyen2019anomaly} proposed unsupervised anomaly detection  by combining a convolutional autoencoder with a U-Net network. The resulting network tries to learn the normal appearance spatial structure through autoecoder and their related motion pattern from optical flow through the second stream which is performed by U-Net. A modified version of inception module has been integrated to the network leading to a patch-based scheme for estimating frame level anomaly. The network has been trained end-to-end using three loss functions: distance loss function and optical flow loss and adversarial loss.

Abate\,\authorcite~\cite{Abati_2019_CVPR} tackled the anomaly detection problem by utilizing the ability of the network on remembering the normal events and evaluating the degree of network surprisal. The remembering aspect of the network is modeled by the reconstruction error of the autoencoder. Also, the suprisal
aspect of the network is modeled by calculating the density of latent features using the autoregressive
network. During the training phase, the joint loss function that combines negative log of the reconstruction error and probability density of latent features is used. The novelty of this work lies in modelling 
the probability density of the latent features using autoregressive model. Most of the autoencoder-based networks are based on element-wise measures such as the squared error. However, the problem of the element-wise metric is its poor performance in modeling the properties of human visual perception. For instance, a small image translation might cause large pixel-wise error~\cite{larsen2015autoencoding}.

The majority of solutions developed for anomaly detection in video are based on unsupervised learning where only normal videos are used for learning and anomaly detection is detected as an outlier detection problem (low probability, anomaly score and reconstruction error). Most video anomaly datasets used for the training and testing are short scenes and cannot generalize to all possible normal patterns. As a result, it is very hard to build a boundary between normal and abnormal events due to the lack of videos that model all possible normal patterns~\cite{chandola2009anomaly}. Sultani\,\authorcite~\cite{sultani2018real} introduced a new approach based on weak supervision where both normal and abnormal videos are used for anomaly detection. In their solution, the anomaly detection problem is formulated within the Multiple Instance Learning (MIL) framework where only bag label is available. In this work, the video is divided into a fixed number of instances (clips) and a multi layer feedfoward perceptron network is trained to predict instance labels based on the deep ranking approach~\cite{sultani2018real}. Recently, Zhong\,\authorcite~\cite{Zhong_2019_CVPR} proposed to address the problem of weak supervision as a supervised learning task under noisy label in which they propose graph convolutional label noise clearner (GCN). This network uses video characteristics such as feature similarity and temporal consistency of video snippets to clean the noise (normal segments of anomalous video). In contrast to its superior performance and since the method is trained using the a whole video at each iteration, the method prone to data correlation~\cite{zaheer2020claws}. To overcome the data correlation problem. Zaheer \authorcite~\cite{zaheer2020claws} proposed to train the network using  batch approach where each batch consist of temporally consecutive segments of a video. In addition, they propose normalcy suppression meachanism to suppress normal features.

The network proposed by Sultani\,\authorcite\,\cite{sultani2018real} deals with instances (clips) independently and does not capture low, intermediate and long temporal information which are very important for video data analysis. Sequence modeling has been used in different fields such as language modeling~\cite{oord2016wavenet},~video summarization~\cite{rochan2018video} and action segmentation~\cite{farha2019ms} to capture the temporal information.

Inspired by the success of temporal convolution in the sequence modelling, we propose a temporal encoding network for anomaly detection in surveillance videos. The proposed network aims to capture the temporal information between video instances. In addition, the problem is also tackled within the MIL framework (weakly supervised) and we formulate a  loss function that uses mean mapping which is smoother than max operation. Also, the loss function penalise the false alarm possibilities which is ultimate desire for real surveillance applications.

\section{Methodology}
\begin{figure*}
	\centering
	\includegraphics[width=\textwidth,scale=0.8]{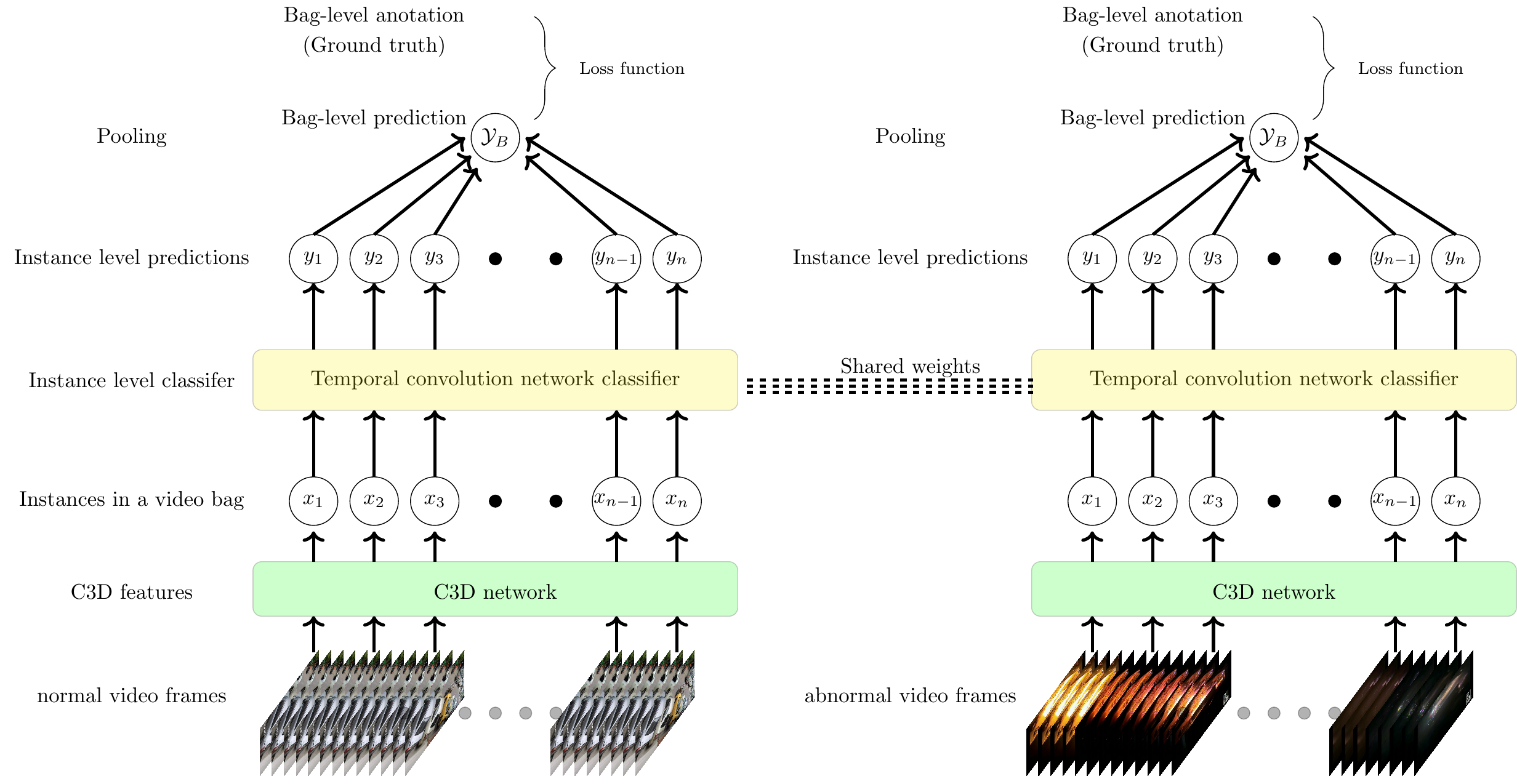}
	\caption{Proposed approach of temporal encoding-decoding network. Normal and abnormal videos are fed into a C3D~\cite{tran2015learning} network to extract spatio-temporal features which are then divided into 32 clips to form instances. These instances are treated as sequential visual information processed by a temporal encoding-decoding network that captures how the features evolve over time and predicts an anomaly score using based on a mapping from instance to label.}
	\label{Fig:proposed approach}
\end{figure*}

The proposed solution is shown in Figure~\ref{Fig:proposed approach}. First, we extract the video spatio-temporal features using C3D network~\cite{tran2015learning}. Then, these feature are divided into a fixed number of non-overlapped clips. These clips form sequential instances in the bag. The proposed temporal encoding-decoding network finds how normal/abnormal feature instances evolve over time. During the training phase, we use both normal and abnormal videos using our a deep ranking loss function to update the network weights.

\subsection{Temporal encoding-decoding network}
Before introducing the network structure, we define the sequence modeling task. Assume that a sequence of features, $\left(x_0,\ldots,x_T\right)$, is given where $x_i\in\mathcal{X}\subset\mathbb{R}^d$ and $d$ is the dimension of the feature space $\mathcal{X}$. At each time step, we want to predict the corresponding output as another sequence $\left(y_0,\ldots,y_T\right)$ of entities in an output space $\mathcal{Y}$. Formally, we can define the sequence modeling network as a mapping function $f:\mathcal{X}^{T+1} \rightarrow \mathcal{Y}^{T+1}$ as follows~\cite{bai2018empirical}:
\begin{eqnarray}
\label{eq:sequence_modeling}
\left(\hat{y}_0,\ldots,\hat{y}_T\right) = f({x}_0,\ldots,{x}_T).
\end{eqnarray}
Note that the number of time steps $T$ is not fixed and depends on the number of sequences of the problem. If the above formulation satisfies the causal constrain, this mean $y_t$ depends on only previous features $(x_0,\ldots,x_{t-1})$ and not on any future inputs. The goal of learning in sequence modeling is to build a network $f$ that minimizes a loss function between actual output and the predicted ones, $\mathcal{L}(y_0,\ldots,y_T,f({x}_0,\ldots,{x}))$. The sequences and the outputs are drawn according to some distribution. In our proposed solution, we deal with video instances as a sequence and the entire learning process as sequence to sequence learning within a weak supervision. In this setting, causal constrain is very important to ensure no information leakage during the training.

The most commonly used network architecture for sequence modeling is the Recurrent Neural Network (RNN)~\cite{goodfellow2016deep}. Recent works have shown that 1-D convolution can also be employed for different tasks that involve sequence modeling such as audio synthesis~\cite{oord2016wavenet}, machine translation~\cite{dauphin2017language} and action segmentation~\cite{lea2017temporal} where no future information is used to predict each output. We have also employed 1D temporal convolution for our network.

Temporal Convolutional Network (TCN) is a network which inherits the properties of convolutional neural networks and uses them to learn a sequence model. TCN is a causal network where there is no information leakage from the future to the past~\cite{bai2018empirical}. Commonly, TCN consists of a 1D fully convolutional network (FCN) and causal convolution in which zero padding of length (kernel size-1) is added to keep the length of the hidden layer the same as the input layer~\cite{bai2018empirical}. 

\begin{figure*}
\centering
	\includegraphics[scale=0.6]{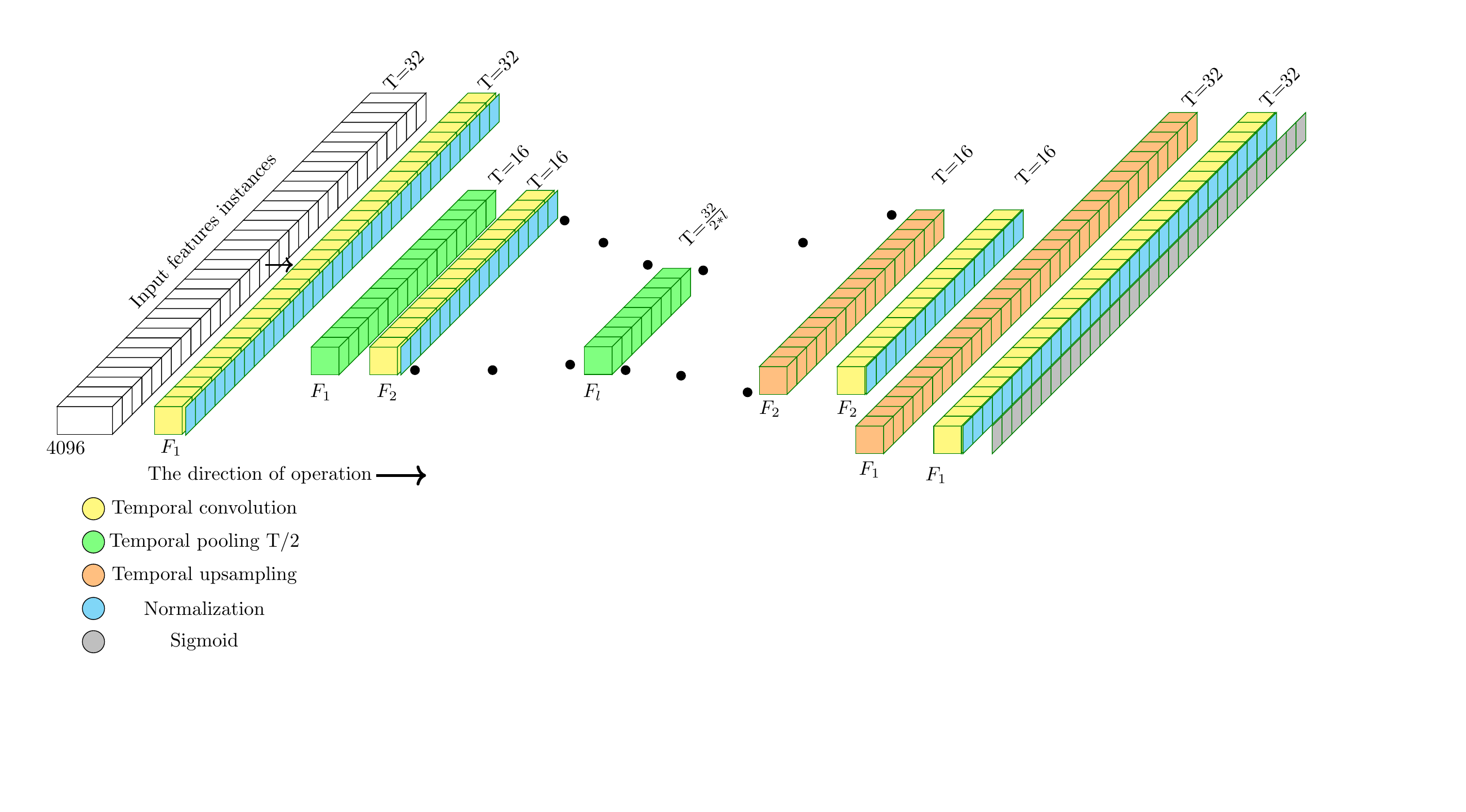}
	\caption{The proposed network, temporal encoding-decoding network. $F_l$ is the number of convolution filters at layer $l$ used in our network. The script $T$ represents the number of instances in temporal domain. The temporal convolution duration length $C_d$ is set to 4 in our network.}
	\label{Fig:proposed network}
\end{figure*}

Inspired by~\cite{lea2017temporal}, we propose a modified TCN network, shown in Figure~\ref{Fig:proposed network}. The new TCN network consists of a temporal encoder/decoder that consists of two steps. Each step has a 1D temporal convolutional layer, a temporal pooling/up-sampling layer, and a channel-wise normalization layer. In particular, the layer $l$ in the encoder/decoder network contains a set of 1D temporal filters, parameterized by tensor $W_{l} \in \mathbb{R}^{F_l\times C_d\times F_{l-1}}$ where $l\in \mathbb{N}\cap[1,L]$ is the layer index, $C_d$ is the temporal convolution duration length and $F_l$ is the number of convolution filters in layer $l$.
These filters are designed to capture the spatio-temporal features and their evolution over the time from one clip to another. The activation function $\hat{E}_{i,t}^{(l)}$ for the $i$-th component ($i\in \mathbb{N}\cap[1,F_l]$) of the $l$-th layer at time step $t$ is defined as:\\
\begin{eqnarray}
\hat{E}_{i,t}^{(l)}=f\Big(b_i^{l}+\sum_{t^{'}=1}^{C_d}\langle W_{i,t^{'}}^{l},E_{0,t+C_d-t^{'}}^{(l-1)}\rangle \Big),
\end{eqnarray}
where $E^{(l-1)}$ is the normalized activation from the previous layer, $f(\cdot)$ is the Leaky Rectified Liner Unit, $b_i^l$ is the bias vector for the $i$-the component in layer $l$, and $\langle\cdot,\cdot\rangle$ is the regular inner product operation. The channel-wise normalization is done as follows~\cite{lea2017temporal}:\\
\begin{equation}
E^{(l)}=1/(m+\epsilon) \hat{E}^{(l)},
\end{equation}
where $\hat{E}^{(l)} = \argmax{i}\hat{E}^{(i,l)}$ is the highest response at time step $t$  and $\epsilon$ is a very small number (usually set as $=1\times 10^{-5}$. Max-pooling with width equal to $T_l = \frac{1}{2}T_{l-1}$ is performed across the temporal domain for each encoder layer.

The decoder architecture is similar to the encoder with the exception of the max-pooling layer which is replayed by the up-sampling layer. The last layer of the decoder is a Sigmoid layer that calculates the anomaly score  with the temporal domain. The training is performed within a weakly supervised framework using normal and abnormal videos. Note that the temporal annotation of abnormal videos is not provided. As a result, the loss function is formulated within the multiple instance learning framework as explained in the following section.

\subsection{Multiple instance deep ranking}

In a multiple instance learning (MIL) context, the task is to learn a classifier based on a set of bags where each bag contains multiple instances. In this setting, the label of the bag is available during training. However, the labels of the instances are not provided. Existing MIL methods can be classified into bag paradigm and instance paradigm~\cite{carbonneau2018multiple}. In the bag paradigm, the aim is to predict the label of the bag, where as, in the instance paradigm, the aim is to predict the instance label~\cite{peng2019address}. Commonly, the main assumption of the MIL is that the bag is positive if at least one instance is positive (for example it is anomaly), otherwise  the bag is negative. This assumption is used to map the label from the instance level to the bag level. However, in most applications such as image segmentation or fine-grained sentiment classification, it is crucial to find the instance label with only bag labels given during training (weakly supervised).

Recently, there has been an increased interest from the computer vision community in studying weakly supervised solutions specially within the MIL framework, in applications such as object detection and localization~\cite{bilen2015weakly,cinbis2016weakly}, image classification~\cite{cabral2014matrix}, and video-based anomaly detection~\cite{sultani2018real}. This is due to the fact that MIL relaxes the need for instance labels (the temporal annotation in our case) and only bag label (video label) is needed. In the following sections, we review MIL deep ranking and related works.

\subsubsection{Mathematical formulation of MIL deep ranking}
Let $\mathbf{X} \in \mathcal{X}$ be the instance-level input random variable and $Y_{\mathbf{X}} \subset \mathcal{Y_{\mathcal{X}}}$ is the instance output, where the space of $\mathbf{X}$ is $\mathcal{X} \subset \mathbb{R}^{d}$, and $d$ is the dimension of the feature vector, also, $\mathcal{Y_{\mathcal{X}}}=\left\lbrace 0,1\right\rbrace $. Assume $\mathbf{B} \in \mathcal{B}$ is the bag-level input random variable and $Y_{\mathbf{B}} \in \mathcal{Y_B}$ is the bag-level output. For instance-based binary classification problem where the training samples are i.i,d, the binary classifier can be defined as follows:\\
\begin{equation}
f(\mathbf{x})={\text{sign}}(g(\mathbf{x})) \in \left\lbrace +1,-1\right\rbrace,
\end{equation}
where $g(\cdot)$ is a mapping function $g:\mathbb{R}^d\rightarrow\mathbb{R}$. 
For the support vector machine (SVM), the optimization problem is reduced to a quadratic programming problem~\cite{suykens1999least}:
\begin{equation}
\begin{aligned}
\min_{\alpha\in \mathbb{R}^d,\beta\in \mathbb{R}} \quad\frac{1}{2} ||\alpha||^2+C\sum_{i=1}^{N}\max \left\lbrace 0,1-y_i(\mathbf{\alpha}^\top\phi(\mathbf{x}_i)+\beta)\right\rbrace,
\end{aligned}
\label{eq:svm}
\end{equation}
where $C>0$ is a penalty parameter, $\beta \in \mathbb{R}$ is a bias parameter, $\alpha \in \mathbb{R}^m$ is an $m$-dimensional classifier weight to be learned, $\mathbf{x}_i \in \mathbb{R}^d$ is the $d$-dimensional instance feature vector, $\phi: \mathbb{R}^d\rightarrow\mathbb{R}^m$ and $y_i$ is the label of the instance. The first term of \eqref{eq:svm} is the $L_2$ regularization and the second term is the hinge loss term, defined as $g(\mathbf{x})=\max(0,1-\mathbf{x})$. It is important to mention that the hinge loss function is not differentiable~\cite{yakhnenko2011multi}. Therefore, different algorithms have been proposed as a solution, such as using the numerical approximation of the hinge loss~\cite{loeff2008scene} or using the generalized hinge loss~\cite{amit2007uncovering}. The maximum margin classifier SVM formulation has been extended to the MIL. In this case, the goal of SVM is to infer the bag label from the instances using the maximum score of the instances in the bag~\cite{andrews2003support}:\\
\begin{equation}
\begin{aligned}
\min_{\alpha\in \mathbb{R}^d,\beta\in \mathbb{R}} \quad\frac{1}{2} ||\alpha||^2+C\sum_{j=1}^{N_B}\max_{\mathbf{x} \in\mathbf{B_j}} \left\lbrace 0,1-\mathcal{Y}_{B_j}(\max(\mathbf{\alpha}^\top\phi(\mathbf{x})+\beta))\right\rbrace,
\end{aligned}
\label{eq:MIL_svm}
\end{equation}
where $N_B$ is the total number of the bags.
Sultani \authorcite~\cite{sultani2018real} reformulate the MIL ranking problem into a rank regression problem. The main assumption is that the anomalous bags should always have a higher anomaly score than the normal bags:
\begin{equation}
\max_{\mathbf{x} \in \mathbf{B}_a}f(\mathbf{x})>\max_{\mathbf{x} \in \mathbf{B}_n}f(\mathbf{x}),\ \ \text{for any } \mathbf{B}_a \in \mathbb{B}_a, \text{ and any } \mathbf{B}_n \in \mathbb{B}_n
\label{eq:mil_ranking}
\end{equation}
where $\mathbb{B}_a$ and $\mathbb{B}_n$ are the given ensembles of abnormal and normal video bags, respectively, and $f(\cdot)$ is the predicted anomaly score for an instance in a bag. The first term of Eq.~\eqref{eq:mil_ranking} represents the instance (segment) that has the highest anomaly score in a given abnormal bag (video), and is highly likely to be an anomaly. However, the second term of Eq.~\eqref{eq:mil_ranking} represents the video segment with the highest anomaly score in a given normal video, which is likely to be a normal instance. The loss equation proposed in~\cite{sultani2018real}, $\mathcal{L}(\cdot)$, includes temporal smoothness of the abnormal video segments as well as sparsity term:
\begin{equation}
\begin{split}
\mathcal{L}(\mathbf{B}_a,\mathbf{B}_n)=\max\left(0,1-\max_{\textbf{x}\in \mathbf{B}_a}f(\textbf{x})+\max_{\textbf{x} \in \mathbf{B}_n}f(\textbf{x})\right) \\
 +\lambda_1 \sum_{\textbf{x}\in \textbf{B}_a} \Delta f(\textbf{x})^2 
+\lambda_2\sum_{\textbf{x}\in\textbf{B}_a}f(\textbf{x}),
\end{split}
\label{eq:sutani_loss}
\end{equation}
where $\lambda_1$ and $\lambda_2$ are hyper-parameters that control the amount of trade-off, and the $\Delta f(\cdot): \mathbb{R}^d\rightarrow [-1,1]$ is the discrete gradient function, defined as
$$
\Delta
f(\textbf{x}_i) 
\triangleq
f(\textbf{x}_i) - f(\textbf{x}_{i+1})
$$
assuming that $\textbf{B}_a = \left(\textbf{x}_1, \cdots , \textbf{x}_n\right)$ where $n$ is the number of instances in the video bag (in our proposed solution examined in the experiments, $n=32$).
The second term in equation~\eqref{eq:sutani_loss} is an $L_2$ regularization term to ensure that the anomaly score of abnormal video varies smoothly from each video segment to the next. The last term in equation~\eqref{eq:sutani_loss} is the $L_1$ regularization sparse that reflects the fact that anomaly in abnormal videos occurs for a short time only. It is evident that the first term of the loss function penalizes the positive bags with low scores.

The problem of the aforementioned loss function is based on the assumption that the bag label is inferred from the maximum score of the instances in the bag, $Y_B=\max(f(x_i))$. The problem arises from the fact that the max function is not smooth~\cite{hu2008multiple} and the optimization suffers from the vanishing gradients~\cite{ilse2018attention}. To alleviate this problem, we propose to use the average difference between normal and abnormal bags in our loss function instead of the max operation:
\begin{equation}
\mathcal{L}(\mathbf{B}_a,\mathbf{B}_n)= 1- \max
\left(
0,d)
\right)  
+
\lambda \sum_{\textbf{x}\in\textbf{B}_a}f(\textbf{x})
\label{eq:our_loss_2}
\end{equation}

where
$$d=\bar{f}(\textbf{B}_a)-\bar{f}(\textbf{B}_n),$$
$$\bar{f}(\textbf{B})= \triangleq {\sum_{\textbf{x}\in\textbf{B}}f(\textbf{x})}\bigg/{|\textbf{B}|},$$

and $|\cdot|$ means ``cardinality of'' or ``the number of elements in''. Indeed, $|\textbf{B}|$ is the bag size for both normal and abnormal video bags that was already denoted by $n$ which is 32 in our experiments.

\begin{figure*}
\centering
	\includegraphics[scale=0.7]{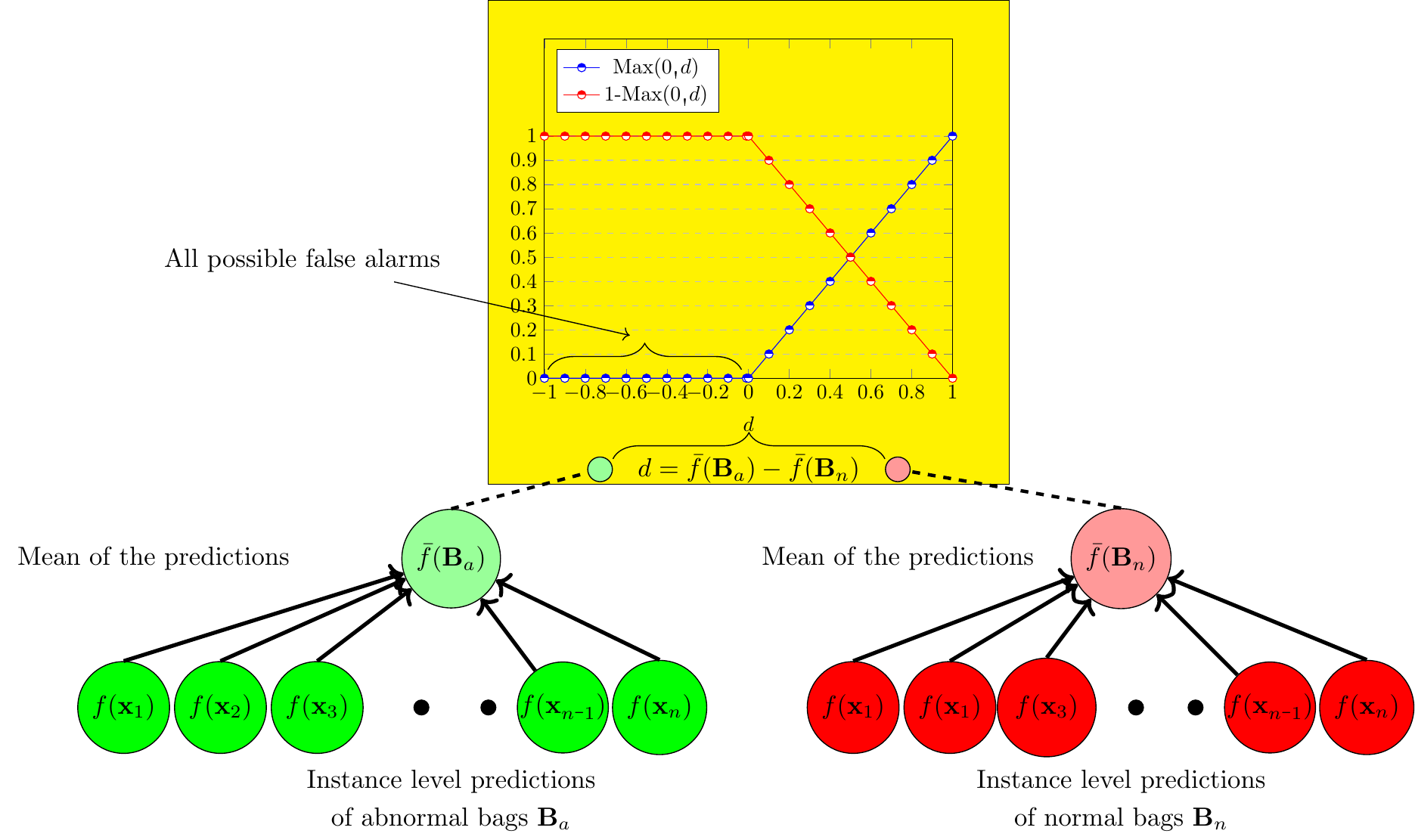}
	\caption{The proposed loss function use the mean distance $d$ between abnormal bag instances score and normal bag instances score. To ensure the abnormal bag always has a higher mean score, we penalize $d$ when it is negative.}
	\label{Fig:proposed loss}
\end{figure*}

This formulation will take into account not only the one instance with the maximum score but also the scores of all instances in the bag.  In addition, the defined loss function maximizes the distance between the average of normal instance score and abnormal instance score, see Figure~\ref{Fig:proposed loss}.
The proposed loss function has only one max operation involved which makes it smoother than that of the one proposed in~\eqref{eq:sutani_loss}. To penalize the loss function of the abnormal bag, similar to~\eqref{eq:sutani_loss}, we added $L_1$ regularization term.\\
%

\section{Experiments}
In this section, we test our proposed temporal encoding-decoding network with the proposed loss function using two public datasets which are UCF-cirme dataset~\cite{sultani2018real} and ShanghaiTech~\cite{luo2017revisit}. We also compare the performance of the proposed solution with the state of the art. In addition, both qualitative and quantitative analysis are carried out.

\subsection{Datasets}
In this paper, we have conducted experiment on two public datasets which are the UCF-crime dataset~\cite{sultani2018real} and the ShanghaiTech~\cite{luo2017revisit}. 
\subsubsection{UCF-Crime} is a large scale dataset of long videos with different scenes that represent real-life situations. The dataset consists of $1900$ videos divided into training sets and testing sets. The training sets consist of 800 normal videos and 810 abnormal videos and the test sets include 150 normal and 140 abnormal videos (290 videos in total). The abnormal videos in both training and testing cover 13 real-world anomalies with following descriptions:~\textit{Abuse},\textit{ Arrest}, \textit{Arson, Assault, Accident, Burglary, Explosion, Fighting, Robbery, Shooting, Stealing, Shoplifting and Vandalism}, see Figure~\ref{Fig:anomaly_examples_1}. The total dataset duration is 128 hours. In this dataset, no temporal (frame-level) annotation is available except for the testing videos. The UCF-crime dataset is the biggest video anomaly dataset and the only one that has multiple scenes with real surveillance videos. Refer to~\cite{sultani2018real} for more details.
\subsubsection{ShanghaiTech} is a medium-scale dataset that contains 437 different videos captured at a university campus. It has 13 different scenes of total 31739 frames of resolution 489$\times$ 856 pixels with different lighting conditions and camera angles. ShanghaiTech dataset is commonly used for unsupervised anomaly detection, thus, there is no abnormal videos for training.  To accommodate this dataset for weakly supervised problem, a new split has been created by Zhong~\authorcite~\cite{Zhong_2019_CVPR} which in this split the training set has normal and abnormal videos. The new split has 175 normal and 63 abnormal training videos and the test set has 155 normal and 44 abnormal videos. For fair comparison, we have used the same split in our experiment.

\subsection{Implementation details}
\subsubsection{Feature extraction and bag generation}:
First, pre-processing operations is performed before feeding it into the C3D network. Each video frame is resized into 240 $\times$ 320 with frame rate fixed to 30 fps. We followed the same technical procedure in~\cite{sultani2018real} to extract the C3D features. We extracted the spatio-temporal features from the fully connected layer (FC6) of the C3D  network~\cite{tran2015learning}. The C3D network computes the C3D features for every 16 frames and then followed by $l_2$ normalization. We divide each video to 32 non-overlapping clips. The video is treated as a bag, and each clip is treated as an instance in the bag. Since we deal with a fixed number of instances per bag, the video instance feature (clip) is generated by taking the average for all 16-frame clip features within that video clip. During the training, a random selection of 30 normal videos and 30 abnormal videos is carried out,  then we feed it as mini-batch to the proposed network as shown in Figure~\ref{Fig:proposed approach}.

\subsubsection{Network implementation}:
The proposed network is implemented using Keras~\cite{chollet2015keras} backend with TensorFlow~\url{(https://www.tensorflow.org/)} and python. We set the temporal convolution kernel length to 4 and different kernel lengths are tested and reported. The number of convolution filters used in our network is set to $\{F_1=512,F_2=128\}$. The last layer of our network is a temporal fully connected layer with a sigmoid activation. We use $l_2$ regularizer for kernel weight parameters for each layer. We used dropout to prevent overfitting. The adaptive subgradient optimizer~\cite{duchi2011adaptive} is used to update network parameters with learning rate set to $0.01$. The $\lambda$ hyper-parameter in our loss function is set to $8\times10^{-5}$ similar to~\cite{sultani2018real}.


\subsection{Metrics} Similar to~\cite{sultani2018real,ionescu2019object,Abati_2019_CVPR}, we used the frame level-based receiver characteristic (ROC) and area under the curve (AUC) metrics to evaluate the proposed method. These metrics are calculated using frame-level ground truth annotation of the test videos. Please note that we do not use equal error rate (EER) since it does not measure anomaly correctly as reported in~\cite{sultani2018real,li2013anomaly}.

\subsection{UCF experimental results}
We compared our proposed method with the baseline method~\cite{sultani2018real} as well as the methods discussed in ~\cite{hasan2016learning,lu2013abnormal}. The proposed method by Lu\,\authorcite~\cite{lu2013abnormal} used a dictionary-based approach to learn the normal pattern from the normal videos and used the reconstruction error as the anomaly score. Hassan\,\authorcite~\cite{hasan2016learning} used deep autoencoder using normal videos to learn normal feature representations and used reconstruction error as an anomaly score.

\begin{figure}
	\centering
	\includegraphics[trim=110 430 150 75,clip,scale=0.7]{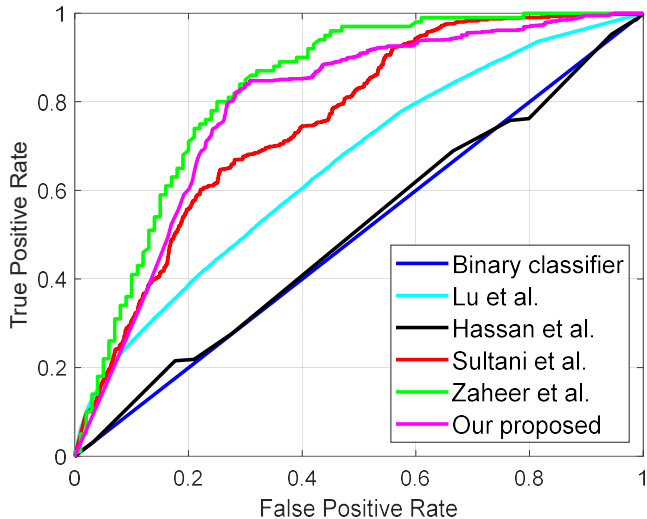}
	\caption{ROC comparative results for different methods on UCF-crime dataset, binary classifier (\textcolor{blue}{blue}), Lu\,\authorcite~\cite{lu2013abnormal} (\textcolor{cyan}{cyan}), Hassan\,\authorcite~\cite{hasan2016learning} (\textcolor{black}{black}), Sultani\,\authorcite~\cite{sultani2018real} (\textcolor{red}{red}), Zaheer \authorcite~\cite{zaheer2020claws} (\textcolor{green}{green}) and our proposed method (\textcolor{magenta}{magenta}).}
	\label{Fig:ROC_curve}
\end{figure}

The quantitative comparisons in terms of ROC and the AUC are reported in Figure~\ref{Fig:ROC_curve} and Table~\ref{Tab:AUC results}. From Figure~\ref{Fig:ROC_curve} it is observed that using normal and abnormal videos in the training phase increases the true positive rate as shown in both Sultani\,\authorcite~\cite{sultani2018real} (red plot) and our proposed method (green plot). In addition, it is evident that our proposed approach has a higher true positive rate compared to base-line method. Table~\ref{Tab:AUC results} shows that our network with the modified loss function achieves the third place compared to the state of the art results for video anomaly detection. We also show that training our network with Sultani\,\authorcite~loss achieves higher results compared to their network because our network exploits the temporal relation between video instances via the spatio-temporal autoencoder. 

\begin{figure*}
	\centering
	\includegraphics[width=\textwidth]{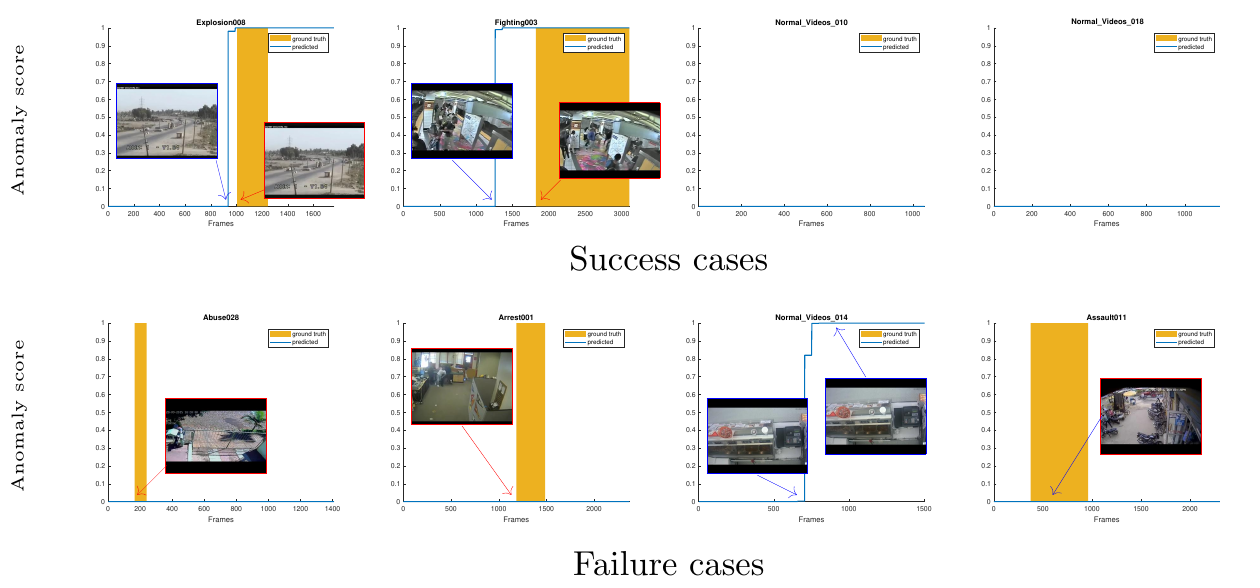}
	\caption{Qualitative results of our method on testing videos. The first row shows an example of success cases and the second row shows an example of failure cases where our method can not produce correct anomaly score. The red frame represents at the ground truth and the blue ones is at the predicted time.}
	\label{fig:qulatitive_results}
\end{figure*}

\begin{figure*}

	\centering
	\includegraphics[width=\textwidth]{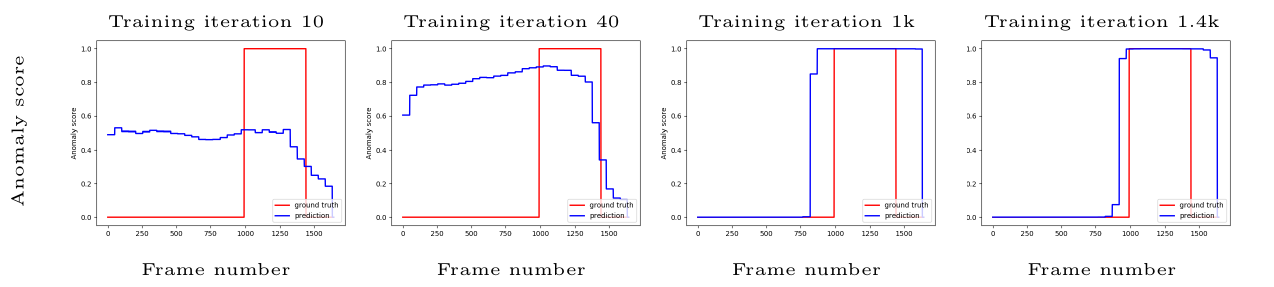}
	\caption{Frame-level anomaly scores evaluation over training iterations generated by our proposed method. Despite of weakly supervised method, our model learns to localize the normal and anomalous regions in the videos by predicting low anomaly score for normal segments and high anomaly score for anomalous segments.}
	\label{Fig:Anomly_score_evalution}
\end{figure*}

Qualitative results on eight different videos that show success and failure cases are reported in Figure~\ref{fig:qulatitive_results}. The first row shows that our method produces a high anomaly score for abnormal videos (explosion and fighting) in a timely manner and generates near-zero anomaly score for normal videos which means that our method generates a low false alarm. We believe that the reason for generating an early anomaly score (first and second columns) is due to the temporal convolution over instances, which proves that our method produces an early detection. The second row shows the cases that our method fails in producing correct anomaly score for different abnormal videos and normal video. 
Also, we provide the evolution of frame-level anomaly scores prediction over several training iterations as shown in Figure~\ref{Fig:Anomly_score_evalution}. It is clear as the number of iteration increases, the proposed method starts predict the correct anomaly scores of both normal and anomalous video segment. 

\begin{table}[ht]
	\label{tab:ROC_curve}
	\centering
	\caption{AUC comparison results of different methods on UCF-crime dataset using C3D features.}	
	\begin{tabular}[t]{lc}
		\toprule
		Method& AUC\%\\ \midrule
		Binary classifier &50.00 \\
		
		Lu\,\authorcite~\cite{lu2013abnormal} &50.60 \\
		
		Hassan\,\authorcite~\cite{hasan2016learning} &65.51 \\
		
		Sultani\,\authorcite~\cite{sultani2018real} & 75.41\\
		Zhong\,\authorcite~\cite{Zhong_2019_CVPR}& 81.08\\
		Zaheer~\authorcite~\cite{zaheer2020claws}&\textbf{83.03}\\
		\midrule
		Our network+ Sultani\,\authorcite~\cite{sultani2018real} loss & 76.41\\
		
		\textbf{Proposed (our network (2 layers)+ our loss)} & 79.49\\
		\bottomrule
	\end{tabular}	
	\label{Tab:AUC results}
\end{table}

\begin{table}
	\centering
	\caption{False alarm rate comparison on normal test videos using C3D features.}	
	\label{Tab:False_alram}
	\vspace{2mm}
	\begin{tabular}{lr}
	    \toprule
	    Method  & False alarm rate \\
	    \midrule
	    Li\,\authorcite~\cite{li2013anomaly} & 27.2\\ 
	    Hasan\,\authorcite~\cite{hasan2016learning} & 3.1 \\
	    Sultani\,\authorcite~\cite{sultani2018real} & 1.9 \\
	    Zhong\,\authorcite~\cite{Zhong_2019_CVPR} & 2.8 \\
	    Zaheer~\authorcite~\cite{zaheer2020claws}&-\\
	    \midrule
	    \textbf{Proposed Method} & \textbf{0.5} \\
	    \bottomrule
	\end{tabular}
\end{table}%

\subsection{False alarm rate}
Similar to~\cite{sultani2018real}, the false alarm rate on normal testing videos is analysed. The reason for this study is that most of the surveillance videos are normal and generating a high false alarm rate is not practical. Therefore, a robust anomaly detection method should report a low false alarm rate on normal videos. We evaluated the performance of our proposed method on normal testing videos. The false alarm rate is reported at 50\% threshold for different methods as shown in Table~\ref{Tab:False_alram}. It is clear that our proposed method has generated a very low false alarm rate in comparison to the based-line method and other methods.

The reason why the proposed approach generates a low false alarm arises due to nature of our loss function. The false alarm occurs when the normal video segment generate high anomaly score. Formally, this happens when $d$ in eq.~\eqref{eq:our_loss_2} is negative, ($(\bar{f}\textbf{B}_n)>\bar{f}(\textbf{B}_b)$), see Figure~(\ref{Fig:proposed loss}). Therefore; our loss function penalise this by setting $d=0$ during the training.
\subsection{ShanghaiTech experimental results}
We trained our proposed model on the ShanghaiTech using the train and test split provided by by Zhong~\authorcite~\cite{Zhong_2019_CVPR}. Since this a recent split, there is no much work reported on this split. We compared our model accuracy with only~\cite{Zhong_2019_CVPR},~\cite{zaheer2020self} and~\cite{zaheer2020claws}. We followed the same protocol in extracting the C3D features and model parameters. We outperform Zhong~\authorcite~\cite{Zhong_2019_CVPR} by significant $11.23\%$ margin and Zhaeer~\authorcite~\cite{zaheer2020self} by $3\%$ margin. However, Zaheer~\authorcite~\cite{zaheer2020claws} outperform our model by $2.36\%$ margin.

\begin{table}[ht]
	\label{tab:ROC_curve_2}
	\centering
	\caption{AUC comparison results of different methods on ShanghaiTech using C3D features.}	
	\begin{tabular}[t]{lc}
		\toprule
		Method & AUC\%\\ \midrule
		Zhong\,\authorcite~\cite{Zhong_2019_CVPR}& 76.44\\
		Zaheer \authorcite~\cite{zaheer2020self}&84.16\\
		Zaheer\,\authorcite~\cite{zaheer2020claws} & \textbf{89.67}\\
\midrule
		\textbf{Proposed (our network+ our loss)} & 87.42
		\\
		\bottomrule
	\end{tabular}	
	\label{Tab:AUC_shang}
\end{table}
\section{Ablative experiments}
In this section, we show how our network architecture is different from the ED-TCN network~\cite{lea2017temporal}. First, we note that the ED-TCN network has been used for action segmentation and trained in a supervised manner~\cite{lea2017temporal} while our network is trained via weak supervision. To accommodate the ED-TCN network to our problem, we changed the last layer, the network regularization and the number of convolution fillers. In addition, we used our loss function and similar parameter settings as mentioned earlier. For the sake of conducting a fair comparison, we set the convolution kernel-size to four for ED-TCN network. Table~\ref{ablative_table} shows that the accuracy (AUC) of ED-TCN network trained by using the proposed loss has lower AUC compared to our network. In addition, we show how our loss function is different compared to Sultani~\authorcite~\cite{sultani2018real} by replacing the max operation by mean operation and report the results which demonstrate the effectiveness of our loss function. 
\begin{table}[t]
	\centering
	\caption{AUC comparison results for different network settings and different temporal networks on UCF cirme dataset.}	
	\begin{tabular}[t]{lc}
		\toprule
		Method & AUC\%\\
		\midrule
		Our loss+ Lea et~al.~\cite{lea2017temporal} network settings & 76.56\\
		Sultani~\authorcite~\cite{sultani2018real} loss+ Lea et~al.~\cite{lea2017temporal} network settings & 78.89 \\ 
		Our loss+ Bi-LSTM~\cite{graves2005bidirectional} network & 50.12\\
		Our network+Sultani et al. loss (with average mapping) &74.53\\
		\midrule
		\textbf{Proposed method(our network+ our loss)} & \textbf{79.49}\\
		\bottomrule
	\end{tabular}
	\label{ablative_table}
\end{table}%

\subsection{Effect of temporal convolution kernel size}
\begin{figure}[t]
	\centering
	\includegraphics[height=6.5cm]{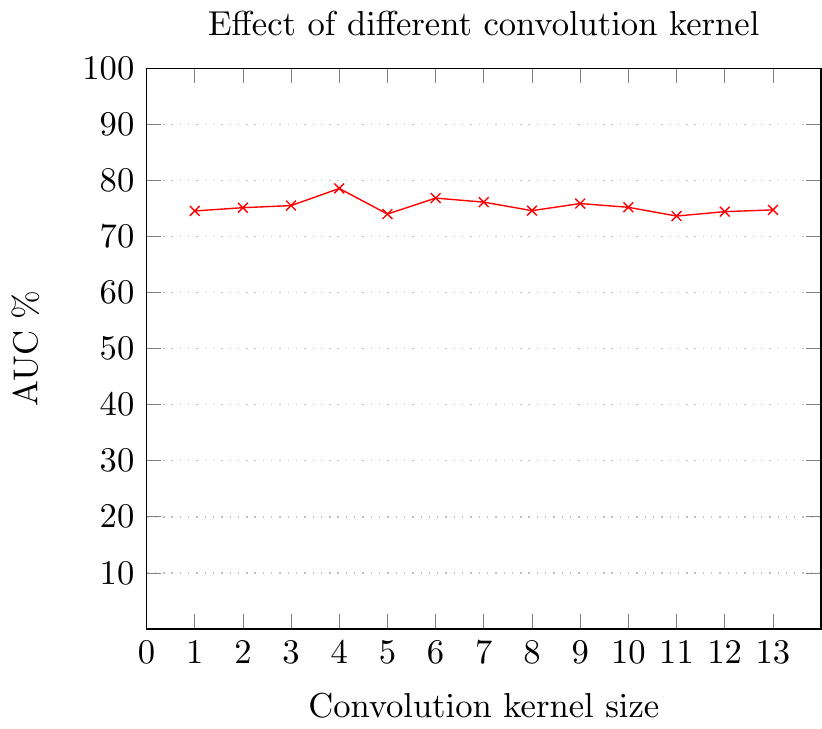}
	\caption{The AUC of the proposed network with different convolution kernel sizes.}
	\label{fig:effect_of_conv_len}
\end{figure}

Figure~\ref{fig:effect_of_conv_len} shows the effect of different convolution kernel sizes on the performance of the proposed approach (in terms of AUC metric). The results demonstrate that temporal convolution with kernel size set to 4 has the highest AUC performance compared to other setups.

\section{Conclusion}
We propose a deep temporal encoding-decoding network for anomaly detection in video surveillance applications. Our proposed solution is based on the deep ranking multiple instance learning where we use normal and abnormal videos during training to localize the anomaly event in real surveillance videos. We deal with video instances (clips) as sequential visual data and build a temporal encoding network that exploits the low, intermediate, and high-level spatio-temporal evolution between the feature instances. Due to the lack of temporal annotation of visual video instances, we use the average sum of the instance predication to pool from the instance-level to bag-level predication. Therefore, our loss function is smoother than using max-pooling in previous work. In addition, the loss function ensure a low false alarm during the training. The results of experiments using normal and abnormal videos in the UCF-crime dataset and ShanghaiTec demonstrate that effectiveness of our proposed solution.

\bibliographystyle{IEEEtran}
\bibliography{mybibfile}


%



\section*{Acknowledgment}
This work was supported by the Australian Research Council through the ARC Linkage Project under Grant LP160101081.

\ifCLASSOPTIONcaptionsoff
  \newpage
\fi



%

%

\begin{IEEEbiographynophoto}{Ammar Kamoona}
received the master's degree in electronic and electrical
engineering from the Swinburne University
of Technology, Melbourne, Australia, in 2016.
He is currently pursuing the Ph.D. degree with
the RMIT University of Technology, Melbourne.
He was an Assistant Lecturer with the Department
of Electrical Engineering, University ofKufa, Iraq,
from 2017 to 2018. His current research interests
include computer vision, RFS filters, robotics and
optimization, and FPGA applications. He was a recipient of two certicates
of excellence in RF circuit design and Stochastic and Survival analysis from
Swinburne University, as well as a Golden Key Certicate for being one of
the top achiever students.
\end{IEEEbiographynophoto}

\begin{IEEEbiographynophoto}{AMIRALI KHODADADIAN GOSTAR}
received the B.Sc. degree in electrical engineering,
the M.Sc. degree in philosophy of science, and
the Ph.D. degree in mechatronics engineering
from RMIT University, where he is currently a
Postdoctoral Research Fellow with the School of
Engineering. His research interests include sensor
management, data fusion, and multitarget tracking.
\end{IEEEbiographynophoto}
\begin{IEEEbiographynophoto}{ALIREZA BAB-HADIASHAR}
received the B.Sc.
and M.Eng. degrees in mechanical engineering
and the Ph.D. degree in robotics from Monash University.
He has held various positions in Monash
University, the Swinburne University of Technology,
and RMIT University, where he is currently a
Professor of mechatronics and leads the Intelligent
Automation Research Group. His main research
interests include intelligent automation in general,
robust data Fitting in machine vision, deep learning
for detection and identication, and robust data segmentation, in particular.
\end{IEEEbiographynophoto}
\begin{IEEEbiographynophoto}{REZA HOSEINNEZHAD}
received the B.Sc.,
M.Sc., and Ph.D. degrees in electrical engineering
from the University of Tehran, Iran, in 1994,
1996, and 2002, respectively. He has held various
positions at the University of Tehran, the Swinburne
University of Technology, The University of
Melbourne, and RMIT University, where he has
worked, since 2010, and is currently a Professor,
and a Research Development Lead, as well as the
Discipline Leader (Manufacturing and Mechatronics)
with the School of Engineering. His main research interests include
statistical information fusion, random Finite sets, multi-object tracking, deep
learning, and robust multi-structure data ftting in computer vision.
\end{IEEEbiographynophoto}




\end{document}